\documentclass[letterpaper, 10 pt, journal, twoside]{IEEEtran}

\usepackage{xcolor}
\usepackage{multirow}
\usepackage{gensymb}
\usepackage{booktabs} 
\usepackage{cite}
\usepackage{algorithm}
\usepackage{algorithmic}
\usepackage{makecell}
\usepackage{textcomp}
\usepackage{array}
\usepackage{graphics} 
\usepackage{epsfig}
\usepackage{times} 
\usepackage{amsmath} 

\usepackage{amssymb}  
\usepackage{xspace}
\usepackage[breaklinks=true,bookmarks=false]{hyperref}
\usepackage[caption=false,font=normalsize,labelfont=sf,textfont=sf]{subfig}
\usepackage{textcomp}
\usepackage{stfloats}
\usepackage{url}
\usepackage{verbatim}
\usepackage{graphicx}
\usepackage{geometry}
\usepackage{cite}

\definecolor{outlier}{rgb}{1, 0.55, 0} 
\definecolor{lidar}{rgb}{0, 0.75, 1} 
\definecolor{radar}{rgb}{1,0,0.75}

\newcommand{\sysname}{\texttt{RaFlow}\xspace}

\newcommand{\highlight}[1]{{\color{black}{#1}}}

\begin{document}

\markboth{IEEE Robotics and Automation Letters. Preprint Version. Accepted June, 2022}{Ding \MakeLowercase{\textit{et al.}}: Self-Supervised Scene Flow Estimation with 4-D Automotive Radar} 

\author{Fangqiang Ding$^{1}$, Zhijun Pan$^{1}$,  Yimin Deng$^{2}$, Jianning Deng$^1$, and Chris Xiaoxuan Lu$^{1}$%
\thanks{Manuscript received: February, 23, 2022; Revised: May, 24, 2022; Accepted: June, 15, 2022.}
\thanks{This paper was recommended for publication by Editor Hyungpil Moon upon evaluation of the Associate Editor and Reviewers' comments. This research is supported by the  EPSRC, as part of the CDT in Robotics and Autonomous Systems at Heriot-Watt University and The University of Edinburgh (EP/S023208/1). (\emph{Corresponding author: Chris Xiaoxuan Lu)}} 
\thanks{$^{1}$Fangqiang Ding, Zhijun Pan, Jianning Deng and Chris Xiaoxuan Lu are with School of Informatics, University of Edinburgh, United Kingdom
        {\tt\footnotesize F.Ding-1@sms.ed.ac.uk}}%
\thanks{$^{2} $Yiming Deng is with the SAIC Motor Technical Center, Shanghai, China.
        {\tt\footnotesize dengyimin@saicmotor.com}}%
\thanks{Digital Object Identifier (DOI): see top of this page.}
}


\title{Self-Supervised Scene Flow Estimation with 4-D Automotive Radar}
\maketitle

\begin{abstract}
Scene flow allows autonomous vehicles to reason about the arbitrary motion of multiple independent objects which is the key to long-term mobile autonomy. While estimating the scene flow from LiDAR has progressed recently, it remains largely unknown how to estimate the scene flow from a 4-D radar - an increasingly popular automotive sensor for its robustness against adverse weather and lighting conditions. Compared with the LiDAR point clouds, radar data are drastically sparser, noisier and in much lower resolution. Annotated datasets for radar scene flow are also in absence and costly to acquire in the real world. These factors jointly pose the radar scene flow estimation as a challenging problem. This work aims to address the above challenges and estimate scene flow from 4-D radar point clouds by leveraging self-supervised learning. A robust scene flow estimation architecture and three novel losses are bespoken designed to cope with intractable radar data. Real-world experimental results validate that our method is able to robustly estimate the radar scene flow in the wild and effectively supports the downstream task of motion segmentation. 
\end{abstract}



\vspace{-1em}
\section{Introduction}

\IEEEPARstart{C}{rucial} to ensuring safe planning and navigation for autonomous vehicles lies in reasoning the motion of dynamic objects in the wild. One representation of such motion is scene flow - a set of displacement vectors between two consecutive frames describing the motion field of a 3D scene. As the 3D extension to optical flow, scene flow intrinsically fits the requirement of the point cloud data produced by a variety of 3D sensors on autonomous vehicles (e.g., LiDAR). Beyond an object-level motion description, scene flow describes the motion field at the point level and gives more fine-grained scene dynamics. Once a scene flow is accurately predicted, an array of applications can be readily enabled, such as dynamic object segmentation, point cloud stitching, and multi-object tracking. 

While recent improvements are witnessed in scene flow estimation by leveraging supervised \cite{wu2020pointpwc,kittenplon2021flowstep3d, gojcic2021weakly} or self-supervised learning \cite{mittal2020just, li2021self, baur2021slim}, these works are mainly dedicated to LiDAR scans and cannot be readily extended to the radar data. Indeed, 4-D millimetre-wave (mmWave) radar as an emerging sensor has been receiving increasing attention from the automotive industry due to a set of complementary advantages over LiDAR. First, cutting-edge radar sensors facilitate richer observations that contain both 3D position and radial relative velocity (RRV) measurement of points in the scene, with the extra velocity measurement naturally benefiting the scene flow estimation. Second, thanks to its operating wavelength in mmWave, these radars are fundamentally robust to adverse weathers, such as fog, rain, dust, or any unfavorable illumination conditions such as darkness, dimness, sun glare etc. Last but not least, radar sensors enjoy more lightweight form factors (i.e., single-chip sizes) and lower cost than LiDAR, lending themselves more affordable to the vehicle platforms with limited budgets or payload, e.g., middle- and low-end vehicles or even delivery robots~\cite{meetscott}. 

\begin{figure}
    \centerline{\includegraphics[scale=0.155]{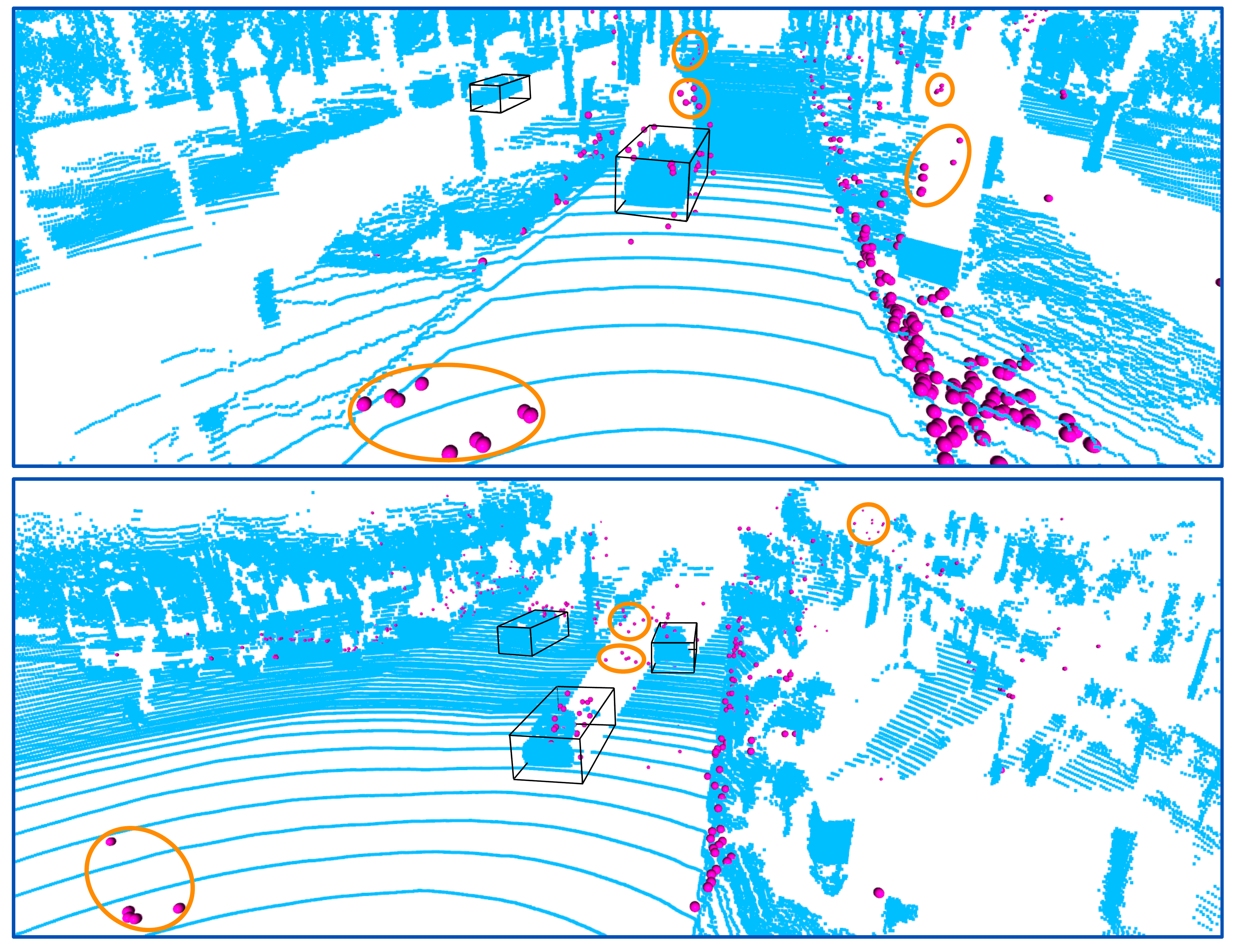}}
    \vspace{-0.25cm}
    \caption{Visualized comparison between LiDAR (\textcolor{lidar}{ blue}) and radar (\textcolor{radar}{magenta}) point clouds. Black bounding boxes come from LiDAR object detection output (highlighted with bold lines). As seen in the figure, radar point clouds are much sparer than LiDAR point clouds with only a small fraction of overlapping points. Moreover, there are plenty of radar noise points, apparent clusters of which are circled in \textcolor{outlier}{orange}. Due to a much lower resolution, many points that should be inside the bounding boxes fall outside. (Best viewed in color) } 
    \label{fig:open}
    \vspace{-2em}
\end{figure}

Despite the aforementioned benefits of 4-D radar, estimating scene flow vectors from radar point clouds is non-trivial and till now, no prior works have been proposed for radar scene flow estimation. 
Compared to the point clouds output by LiDAR, radar point clouds are significantly sparser, noisier and in much lower resolution (c.f. Fig.~\ref{fig:open}). Concretely, a single sweep of 4-D radar point cloud only has several hundreds of points ($< 1 \%$ of a common LiDAR point cloud). This sparse nature hinders the robustness of local feature extraction methods. Moreover, radar returns are usually corrupted by noise due to the multi-path effect and specular reflection. As a result, a non-negligible amount (up to $20\%$ in our case) of radar points are outliers, undermining the point association step of scene flow estimation. Meanwhile, the distance and angular measurement resolution of mmWave radar are constrained by the limited number of antennas for cost reasons on chip. For instance, the angular resolution of our radar is an order of magnitude worse than that of LiDAR. The above challenges are exacerbated by the fact that there are no public 4-D radar datasets and point-level scene flow annotations are costly to acquire. This fact unfortunately limits the usability of many supervised learning-based methods proposed recently~\cite{kittenplon2021flowstep3d, gojcic2021weakly,gu2019hplflownet,behl2019pointflownet,wang2020flownet3d++,wu2020pointpwc,puy2020flot, zuanazzi2020adversarial}.

To address the above challenges and unleash the potential of 4-D radar, we propose a novel self-supervised learning approach, termed \sysname, for robust scene flow estimation. \sysname~starts with a module bespoke to radar and derives a basic scene flow between two radar point clouds. \sysname then utilizes the RRV measurements to generate a static mask and obtain the rigid ego-motion through the Kabsch algorithm~\cite{kabsch1976solution}. The points identified as static will be further refined from rigid flow vectors. The overall architecture can be trained end-to-end. To enable robust self-supervised learning, we design three task-specific losses to cope with sparse and noisy radar point clouds. Without the need of labels, our proposed loss functions can collectively regularize the model to learn to estimate scene flow by exploiting the underlying supervision signals embedded in the radar measurements. For evaluation, we drove a vehicle equipped with radar and other sensors for 43.6km in the wild and collected a multi-modal dataset. Real-world evaluation results demonstrate that, compared with  state-of-the-art scene flow methods designed for LiDARs, \sysname is a more suitable approach for the 4-D radar and can yield much more robust scene flow estimation. 
\highlight{
To summarize, our contributions are as follows:
\begin{itemize}
\item A first-of-its-kind work to investigate the scene flow estimation from 4-D radar data. A self-supervised learning framework is proposed for scene flow estimation without resorting to costly annotations.
\item The proposed \sysname is capable of handling sparse, noisy, and low-resolution radar point clouds and can effectively exploit the unique RRV measurement of radars to drastically strengthen the estimation robustness.
\item We drove a vehicle in the wild to collect a multi-modal dataset for systematic evaluation. Experimental results demonstrate the superior performance of \sysname compared with state-of-the-art point-based scene flow estimation methods. Our source code will be released to the community at  \href{https://github.com/Toytiny/RaFlow}{\emph{https://github.com/Toytiny/RaFlow}}.
\end{itemize}
}
\vspace{-1em}
\section{Related Works}

\subsection{Supervised scene flow estimation on point clouds}

Till present, supervised learning is still the dominant approach to scene flow predictions esp. when it is used together with deep neural network (DNN)~\cite{liu2019flownet3d,gu2019hplflownet,behl2019pointflownet,wang2020flownet3d++,wu2020pointpwc,li2021hcrf,puy2020flot, zuanazzi2020adversarial}. 
Early efforts in this vein \cite{liu2019flownet3d, gu2019hplflownet} designed DNN architectures based on the layers~\cite{qi2017pointnet++, su2018splatnet} to process unordered point cloud inputs for effective scene flow estimation. Concurrently, a multi-task pipeline \cite{behl2019pointflownet} was proposed to predict scene flow and detect objects simultaneously with voxel-based networks. Subsequent works improved scene flow estimation accuracy by incorporating geometric constraints~\cite{wang2020flownet3d++}, optimal transport~\cite{puy2020flot}, cost volume~\cite{wu2020pointpwc}, Con-HCRF~\cite{li2021hcrf} and adversarial learning~\cite{zuanazzi2020adversarial}. 
Nevertheless, supervised DNN methods come with the demand of large-scale annotated scene flow data for training, which are costly to acquire in practice. As a cost-effective alternative, synthetic \cite{mayer2016large} or converted~\cite{menze2018object} datasets are used for offline DNN training. However, these methods fall short of the emerging 4-D radar as an accurate radar synthesizer/generator itself is an unsolved topic, not to mention the common discrepancy between any simulator and real-world data. In~\cite{gojcic2021weakly}, a weakly-supervised pipeline was proposed to lessen the direct requirement of dense scene flow labels at the cost of demanding annotated background mask which still needs significant human effort to carefully inspect the sparse and non-intuitive radar point clouds.

\begin{figure}[t]
    \centerline{\includegraphics[scale=0.255]{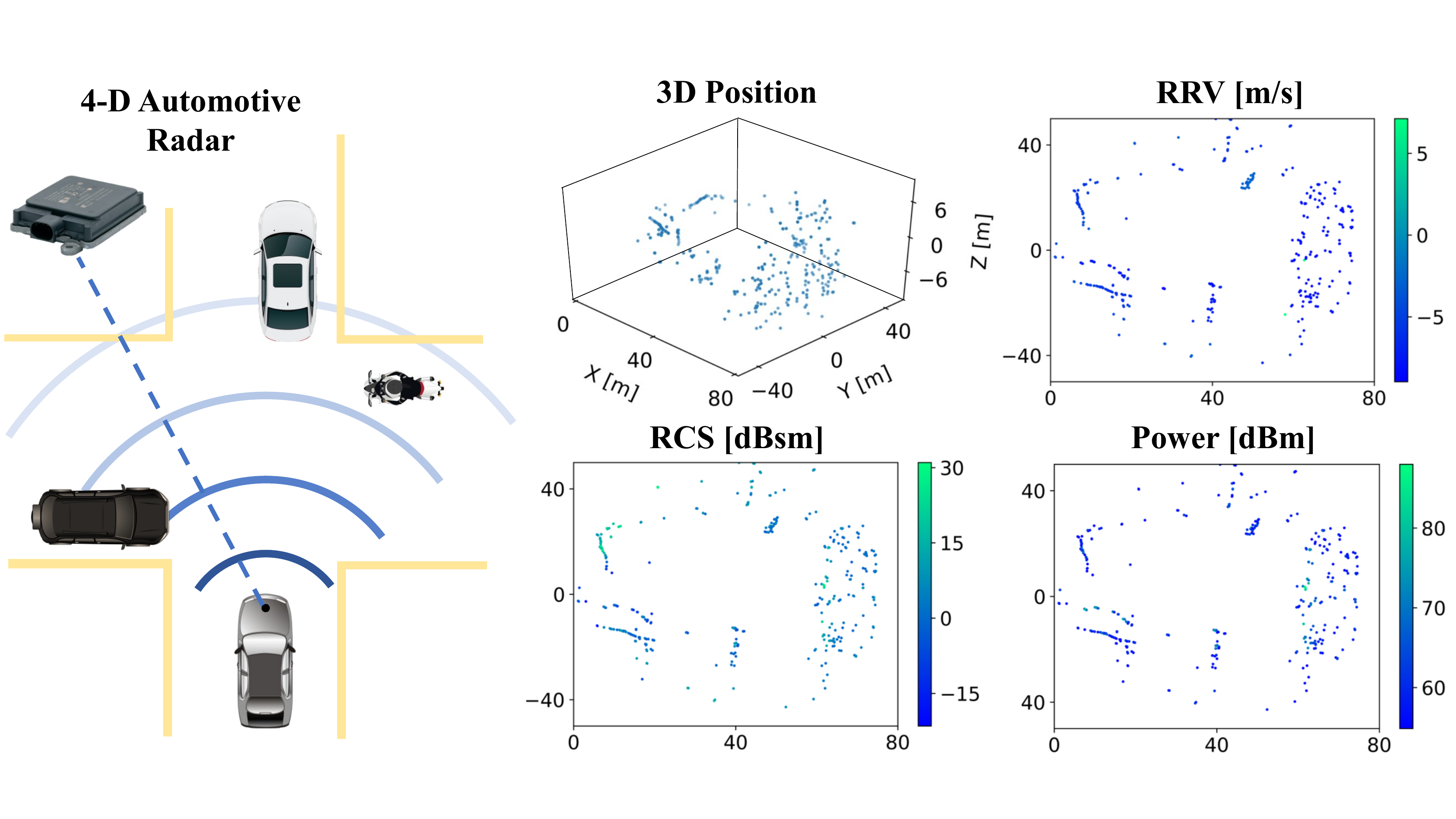}}
    \vspace{-1em}
    \caption{Visualization of the 6-dim measurements of a radar point cloud, including the 3D positional information, RRV, RCS and Power measurements. \highlight{The RRV value range here is -85$\sim$45 m/s. The RCS of a target is the hypothetical area that intercepts the radar's transmitted signals and reflects them back to the radar receiver, whose unit is usually decibels relative to a square meter (dBsm). The Power measurement equals the signal power level received from a target, whose unit is usually decibels relative to one milliwatt (dBm)~\cite{barton2004radar}}. Note that the latter three features are also 3D yet we visualise them in bird's-eye view and color them by their values per point for readability. }
    \label{input}
    \vspace{-1.5em}
\end{figure}

\vspace{-1em}
\subsection{Self-supervised scene flow estimation on point clouds}
There is a recent surge on self-supervised learning based methods that train a scene flow estimator without the need of labelled data~\cite{mittal2020just,wu2020pointpwc, pontes2020scene,tishchenko2020self,kittenplon2021flowstep3d,baur2021slim,li2021self}. 
These methods often exploit or mine supervision signal from the input itself by either designing specific losses~\cite{mittal2020just,wu2020pointpwc,baur2021slim} or generating pseudo scene flow labels~\cite{li2021self} to guide the model training. Yet, all these methods only consider dense LiDAR point clouds as input and their performance may degrade greatly when applied to radars in the following aspects.

\noindent\textbf{Temporal matching.} Most self-supervised methods~\cite{mittal2020just,wu2020pointpwc, pontes2020scene,tishchenko2020self,kittenplon2021flowstep3d,baur2021slim} rely on temporal directed or bidirectional matching between two point clouds to exploit pseudo correspondence information for training. This temporal matching scheme works well for LiDAR point clouds since most points have their close neighbours in the next frame to be matched with. However, it cannot be transferred to radar point clouds directly because: a) the sparsity and noise of radar data make temporal matching between consecutive frames difficult; b) radar points have a much lower resolution which hampers accurate matching between correspondences. 
In this work, we propose a radar-friendly soft Chamfer loss to mitigate the negative effects of the above issues resulting from the inherent properties of the radar sensor. 

\noindent\textbf{Rigid ego-motion estimation.} Scene flow can be respectively disentangled into two parts induced by rigid ego-motion and moving objects. Separating these two components can not only allow us to obtain sensor ego-motion but also enable holistic refinement of flow vectors for static points. Some works~\cite{behl2019pointflownet,gojcic2021weakly} jointly estimate ego-motion, scene flow and object rigid motion but need various labels for supervision. Method in~\cite{tishchenko2020self} regresses ego-motion in the first stage and then estimates the non-rigid object motion with the transformed point cloud. A recent work~\cite{baur2021slim} predicted point-level motion classification with the network and solved the ego-motion using the differentiable Kabsch algorithm~\cite{kabsch1976solution}. However, these methods are either susceptible to the non-negligible amounts of outliers in radar point clouds~\cite{tishchenko2020self}, or highly rely on a large amount of artificial labels~\cite{baur2021slim} that are hard to access with the sparser and noisier radar data. 
Our static flow refinement module also use the Kabsch algorithm to solve ego-motion but segment static points with respect to the RRV measurements instead of network output. 


\vspace{-1em}
\section{Proposed Method}
\subsection{Overview}

\begin{figure}[t]
    \centerline{\includegraphics[scale=0.25]{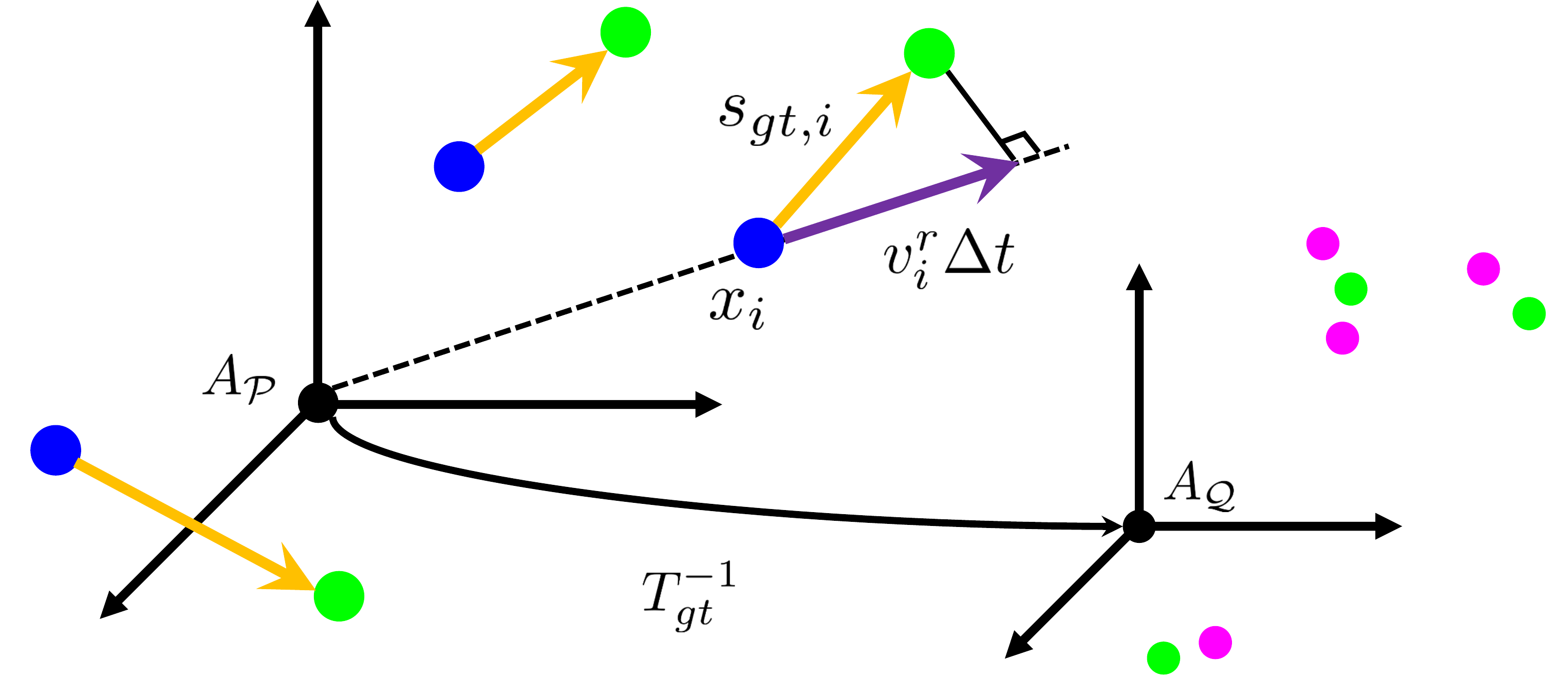}}
    \caption{Illustration of radar scene flow and 4-D radar RRV measurement. Points from point cloud $\mathcal{P}$ and $\mathcal{Q}$ are colored as {\color[rgb]{0,0,1} blue} and {\color[rgb]{1,0,1} magenta}, respectively. {\color[rgb]{0,1,0} Green} is used to denote points warped by the truth flow vectors. In above figure, we assume that all points keep constant velocity during $\Delta t$. $A_{\mathcal{P}}$ and $A_{\mathcal{Q}}$ indicate two local sensor coordinates where point cloud $\mathcal{P}$ and $\mathcal{Q}$ are captured respectively. $T_{gt}^{-1}$ is the true ego-motion. Note that the point clouds in these consecutive frames are non-bijective, as the point gap shown in the $A_{\mathcal{Q}}$ coordinate. (Best viewed in color)}
    \label{fig:rrv}
    \vspace{-1.5em}
\end{figure}

\noindent \textbf{Problem Formulation.}
Given two consecutive point clouds  $\mathcal{P}\in\mathbb{R}^{N_1\times C}=\{\{x_i,f_i\}\in\mathbb{R}^{C}\}_{i=1}^{N_1}$ and $\mathcal{Q}\in\mathbb{R}^{N_2\times C}=\{\{y_i,g_i\}\in\mathbb{R}^C\}_{i=1}^{N_2}$ captured by the same radar, the scene flow estimation aims to derive a set of 3D vectors  $\mathcal{S}\in\mathbb{R}^{N_1\times 3}=\{s_{i}\in\mathbb{R}^3\}_{i=1}^{N_1}$ that represents the displacement of each point $x_i\in\mathcal{P}$ to its corresponding position ${x}'_i=x_i+s_i$ in the scene described by point cloud $\mathcal{Q}$. \highlight{Here, $N_1$, $N_2$ denote the number of points and $C$ is number of input channels.}
Unlike traditional point clouds that only include the 3D positional information for each point (i.e., $x_i,y_i\in\mathbb{R}^3$), the radar point clouds in this work have another 3D features (i.e., $f_i,g_i\in\mathbb{R}^3$) including radial relative velocity (RRV) measurement, Radar Cross Section (RCS) and Power measurement in conjunction with the 3D positional information. RRV characterises the instant motion level of the objects in the scene relative to the ego-vehicle while RCS and Power measurement characterise reflectivity of those objects in different aspects~\cite{barton2004radar}, as shown in Fig.~\ref{input}. All of them come directly from the radar scans and complement the original 3D positional information by providing more scene semantics and motion clues. In alignment with the real-world radar data, here we do not assume absolute bijective mapping between two point clouds, and $\mathcal{Q}$ thus do not necessarily contain the corresponding point ${x}'_i$ as seen in Fig.~\ref{fig:rrv}.

\begin{table}[t]
    \renewcommand\arraystretch{1}
    \setlength\tabcolsep{3pt}
    \centering
    \caption{Architecture spec of our ROFE module.}
    \label{table:arch}
    \begin{tabular}{cccccc}
    \toprule
       \bf{Component}&\bf{Layer type}&\bf{Radius}&\bf{Samples}&\bf{MLP width}\\
    \midrule
       \multirow{4}*{{multi-scale encoder}}
       & set conv & 2.0 & 4 & [32, 32, 64]\\
       & set conv & 4.0 & 8 & [32, 32, 64]\\
       & set conv & 8.0 & 16 & [32, 32, 64]\\
       & set conv & 16.0 & 32 & [32, 32, 64]\\
    \midrule
       {cost volume layer}  
       & - & - & 8 & [512, 512, 512]\\
    \midrule
       \multirow{4}*{{flow decoder}}
       & set conv & 2.0 & 4 & [512, 256, 64] \\
       & set conv & 4.0 & 8 & [512, 256, 64] \\
       & set conv & 8.0 & 16 & [512, 256, 64] \\
       & set conv & 16.0 & 32 & [512, 256, 64] \\
       & output layer & - & - & [256, 128, 64, 3] \\
    \bottomrule
    \end{tabular}
    \vspace{-1.5em}
\end{table}

\noindent \textbf{Overall Architecture.} 
Fig.~\ref{pipeline} illustrates the overall architecture of \sysname, which is composed of a \emph{Radar-Oriented Flow Estimation} (ROFE) module and a \emph{Static Flow Refinement} (SFR) module. The ROFE module takes point clouds $\mathcal{P}$ and $\mathcal{Q}$ as the input pair and firstly estimates a coarse scene flow $\mathcal{S}_c\in\mathbb{R}^{N_1\times 3}=\{s_{c,i}\in\mathbb{R}^3\}_{i=1}^{N_1}$, where each flow vector is unconstrained. Based on the coarse scene flow, the SFR module generates a static mask $\mathcal{M}_{s}$ using Algorithm~\ref{alg:mask} and then estimates a global rigid transformation $T_r\in SE(3)$ with the differentiable Kabsch algorithm~\cite{kabsch1976solution}. The rigid scene flow estimation $\mathcal{S}_r\in\mathbb{R}^{N_1\times 3}=\{s_{r,i}\in\mathbb{R}^3\}_{i=1}^{N_1}$ can be reparameterized through ${s}_{r,i}=(T_r-I_4)\tilde{{x}}_i$, where $\tilde{x}_i$ denotes $x_i$ in the homogeneous coordinate.
Finally, we refine the coarse flow vectors of all static points with the rigid scene flow $\mathcal{S}_r$ to obtain the final scene flow $\mathcal{S}$. The entire architecture is end-to-end trainable.
With this architecture, we can not only predict a refined scene flow $\mathcal{S}$ but also address the motion segmentation and rigid ego-motion estimation problems. The static mask $\mathcal{M}_s$ can be used to segment moving points, while the rigid transformation $T_r$ describes the ego-motion of radar. 

\vspace{-1em}
\subsection{Radar-Oriented Flow Estimation (ROFE) Module}
\begin{figure}[t]
    \centerline{\includegraphics[scale=0.1]{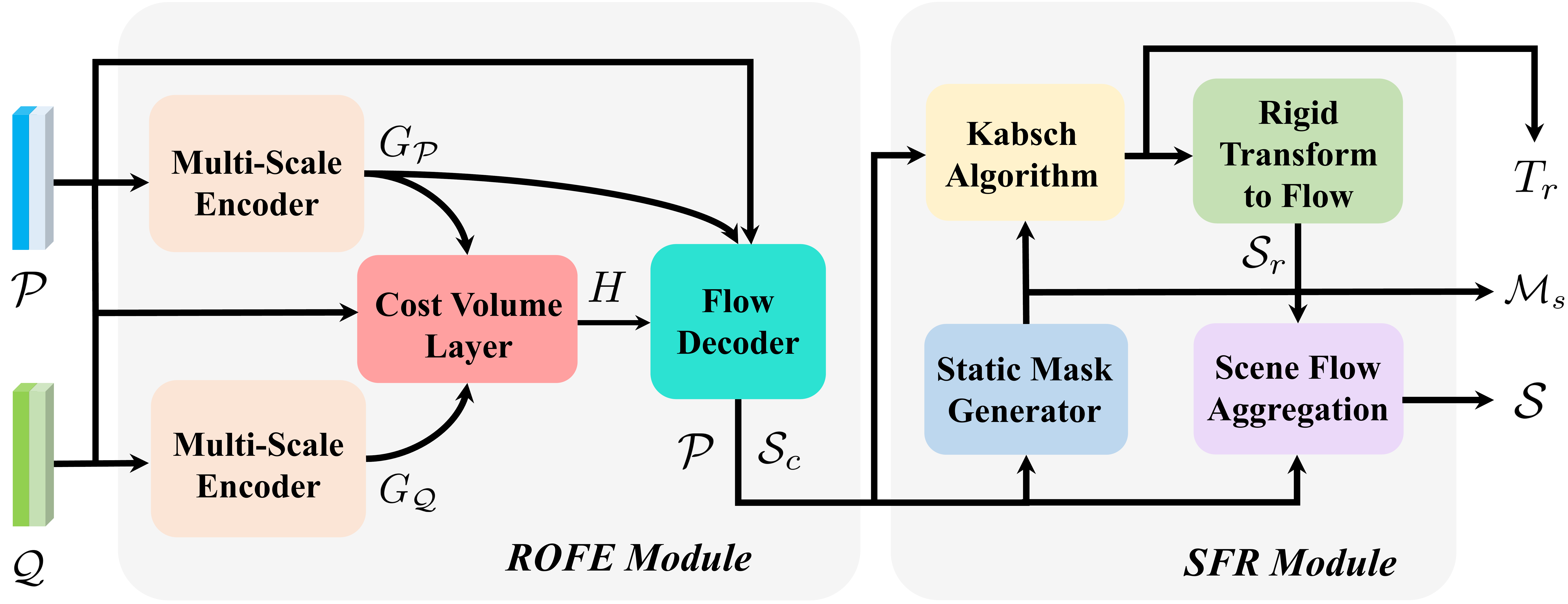}}
    \caption{Overview of our radar scene flow estimation pipeline. Two point clouds $\mathcal{P}$ and $\mathcal{Q}$ are used as the input to \sysname. The output includes the aggregated final scene flow $\mathcal{S}$, the static mask $\mathcal{M}_s$ and the rigid transformation $T_r$. Note that the two multi-scale encoders share the same weights. The entire architecture is end-to-end trainable.} 
    \label{pipeline}
\end{figure}

As shown in Fig.~\ref{pipeline}, the ROFE module consists of three components: multi-scale encoder, \highlight{cost volume layer} and flow estimator, which are explained below. 
\highlight{The detailed layer parameters of the ROFE module can be seen in Table~\ref{table:arch}}.

\noindent\textbf{Multi-scale Encoder.} 
Extracting robust local features from radar points clouds is hindered by two major factors: a) the severe sparse nature of radar, b) uneven point density. While enlarging the receptive field can mitigate the sparse issue, it is still hard to tackle uneven point density with single-scale feature extraction. Inspired by the discussion in~\cite{qi2017pointnet++}, we adopt a multi-scale grouping scheme to encode features on radar point clouds. Given input point set $\mathcal{P}$, we use multiple \emph{set conv} layers~\cite{liu2019flownet3d} to group multi-scale local features with different radius neighbourhoods specified by a radius set $R$. To include global knowledge, we also use a channel-wise max-pooling operation to aggregate global features following~\cite{qi2017pointnet} and concatenate it to per-point local features. As a result, we can obtain local-global features ${G}_\mathcal{P}\in\mathbb{R}^{N_1\times(2\sum_{k=1}^{N_{sc}}{C_k})}$ for point cloud $\mathcal{P}$ and $G_\mathcal{Q}\in\mathbb{R}^{N_2\times(2\sum_{k=1}^{N_{sc}}{C_k})}$ for $\mathcal{Q}$ similarly. 

\noindent\textbf{Cost Volume Layer.} \highlight{To robustly and stably correlate features across two frames, we adopt the \emph{Cost Volume} layer proposed by~\cite{wu2020pointpwc}.} This cost volume layer aggregates costs in a patch-to-patch manner, which is particularly useful to mitigate the low-resolution issue of radar point clouds and the resultant non-bijective mapping across frames. After passing the point clouds and local-global features through \highlight{cost volume layer}, we can attain correlated features $H\in\mathbb{R}^{N_1\times C_{cor}}$. 

\noindent\textbf{Flow Decoder.}
\highlight{To decode the flow estimation from features, we firstly form flow embedding by concatenating correlated, local-global and input features of point cloud $\mathcal{P}$.}
With flow embedding and point cloud $\mathcal{P}$, we use another multi-scale encoder to group embedding features to include spatial smoothness for the final output. We denote these features as $U\in\mathbb{R}^{N_1\times (2\sum_{k=1}^{N_{sc}}{C'_k})}$ from multiple \emph{set conv} layers. Lastly, $U$ is fed into the final four-layer MLP whose output is the coarse scene flow $\mathcal{S}_c$.

\vspace{-1.5em}
\subsection{RRV Measurement and Scene Flow}\label{rrv}

One important property of 4-D radars is that their RRV measurements from Doppler effects are intuitive self-supervision signals for scene flow estimation. RRV measurements describe the moving speed of ambient objects relative to the observer in the radial direction. 
We denote the RRV measurements of point cloud $\mathcal{P}$ as $\mathcal{V}_\mathcal{P}^r \in \mathbb{R}^{N_1} = \{v_i^r\in\mathbb{R}\}_{i=1}^{N_1}$, where $v_i^r$ is positive when point $x_i$ is moving away from the observation point. 
Generally, $\mathcal{V}^{r}_{\mathcal{P}}$ can be used as a component of the input features to include point-level motion cues. However, we argue that RRV measurement has the potential to play more roles in \sysname. 
Assuming the velocity of point $x_i$ keeps constant during time interval $\Delta t$ between two scans, we can reach the following equation:
\begin{equation}\label{velo}
v_i^r \Delta t={s}_{gt,i}^\top\frac{x_i}{||x_i||}
\end{equation} 
where ${s}_{gt,i}$ is the true flow vector of point $x_i$. Eq.~\ref{velo} means that the projection of flow vector on the radial direction equals the measured RRV times $\Delta t$, as explained in Fig.~\ref{fig:rrv}. The constant velocity assumption is rational because the time interval between two radar scans is usually very short (e.g., 100ms) 
so that the average velocity of points can be seen as an approximation of the instantaneous velocity \highlight{in most cases}. As we will see in the rest of this paper, RRV is crucial for our self-supervised learning framework even without the availability of the truth scene flow. We propose a static mask generation module in Section~\ref{refinement} and formulate a radial replacement loss function in Section~\ref{loss}, which are both inspired by Eq.~\ref{velo}.

\setlength{\textfloatsep}{0.1cm}
\begin{algorithm}[t]
\caption{Static Mask Generation}
\begin{algorithmic}[1] 
    \REQUIRE Point cloud $\mathcal{P}$, coarse scene flow $\mathcal{S}_c$, RRV measurements $\mathcal{V}_{\mathcal{P}}^r$ of point cloud $\mathcal{P}$, time interval $\Delta t$ between point cloud $\mathcal{P}$ and $\mathcal{Q}$, preset threshold $\zeta$
    \ENSURE Static mask $\mathcal{M}_s=\{m_{s,i}\in\{0,1\}\}_{i=1}^{N_1}$
   \STATE Get the warped point cloud $\mathcal{P}'_c =\{\{x'_{c,i},f_i\}\in\mathbb{R}^{C}\}_{i=1}^{N_1}$ with $\mathcal{S}_c$ and $\mathcal{P}$ by: ${x}'_{c,i}\leftarrow{x_i}+{s}_{c,i}$
\STATE Form correspondences $\{x_i,x'_{c,i}\}_{i=1}^{N_1}$ between $\mathcal{P}$ and $\mathcal{P}'_c$
\STATE Feed $\{x_i,x'_{c,i}\}_{i=1}^{N_1}$ into Kabsch algorithm and get $T_{cr}$
\STATE Compute $\mathcal{S}_{cr}\in\mathbb{R}^{N_1\times 3}=\{s_{cr,i}\in\mathbb{R}^3\}_{i=1}^{N_1}$ with $T_{cr}$ by $s_{cr,i}\leftarrow (T_{cr}-I_4)\tilde{{x}}_i$
\FOR{$i=1$ to $N_1$}
\STATE Obtain $s_{cr,i}$'s radial component: $s_{cr,i}^r\leftarrow s_{cr,i}^{\top}\frac{x_i}{||x_i||}$
\STATE Derive the radial shift residual: $r_i\leftarrow s^{r}_{cr,i}-v_i^{r}\Delta t$ 
\STATE Compute the relative residual: $e_i\leftarrow|\frac{r_i}{v^r_i\Delta t}|$
\IF{$e_i\leq\zeta$}
\STATE {$m_{s,i}\leftarrow 1$}
\ELSE
\STATE $m_{s,i}\leftarrow 0$
\ENDIF
\ENDFOR
\end{algorithmic}
\label{alg:mask}
\end{algorithm}
\setlength{\floatsep}{0.1cm}

\vspace{-1.5em}
\subsection{Static Flow Refinement (SFR) Module }\label{refinement}

Due to the intrinsic sparsity and non-negligible noise of radar data, the scene flow $\mathcal{S}_c$ estimated from the ROFE module is only coarse-grained and needs to be refined further. 
Since the motion of stationary scene points is caused by the radar ego-motion, one plausible refinement is to regularize those flow vectors of all static (background) points via a rigid transformation matrix respective to radar ego-motion. Unfortunately, \emph{accurate} ego-motion is often in absence in the real world or costly to acquire in practice. \sysname~to circumvent this ego-motion absence, we first propose a static mask generator based on the RRV measurements available in a radar scan and then leverage the Kabsch algorithm~\cite{kabsch1976solution} to obtain the rigid transformation of the identified static points. \highlight{Particularly, similar to other scene flow works in this vein (e.g.,~\cite{baur2021slim,gojcic2021weakly}), we adopt a specialized form} \highlight{of the Kabsch algorithm, where the centered point coordinates of two sets of paired points are firstly computed by subtracting the centroid coordinates. Then, the optimal rotation matrix is solved through a singular value decomposition. The translation vector is finally restored by comparing two centroid coordinates after using the rotation matrix for compensation. } 

Concrete details of our proposed static mask generation algorithm can be seen in Algorithm~\ref{alg:mask}. Based on our observation that most points in the scene are stationary, we intuitively assume that all points are stationary and then utilize the Kabsch algorithm~\cite{kabsch1976solution} to get an intermediate transformation $T_{cr}\in SE(3)$. Obviously, ${T}_{cr}$ is only a coarse transformation as not all points in the scene are stationary. With ${T}_{cr}$, an intermediate rigid scene flow $\mathcal{S}_{cr}$ can be computed accordingly. With Eq.~\ref{velo} in place, we judge if points are static or moving by comparing their relative residual $e_i$ with a preset threshold $\zeta$. Here, $e_i$ is defined as the difference between the radial displacement induced by the intermediate rigid flow vector $s_{cr,i}$ and RRV measurement $v_{i}^r$.


After applying the static mask $\mathcal{M}_s$ on point cloud $\mathcal{P}$ and the warped one $\mathcal{P}'_{c}$, we feed only static correspondences into the Kabsch algorithm again. \highlight{As exhibited in Fig.~\ref{pipeline}, a more reliable rigid-motion transformation ${T}_r$ can be thus obtained. Then, we derive the rigid scene flow $\mathcal{S}_r$ from $T_r$, which can be shown as the \emph{Rigid Transform to Flow} block in Fig.~\ref{pipeline}.} Lastly, we aggregate the final scene flow $\mathcal{S}$ by:
\begin{equation}\label{rigid}
    {s}_{i}=\left\{
    \begin{aligned}
    s_{r,i}~~\mathrm{if}~m_{s,i}=1\\
    s_{c,i}~~\mathrm{if}~m_{s,i}=0
    \end{aligned}
\right.
\end{equation}

\subsection{Loss Function}\label{loss}

We leverage three types of self-supervision signals for model training: radial displacement loss $\mathcal{L}_{rd}$, soft Chamfer loss $\mathcal{L}_{sc}$ and spatial smoothness loss $\mathcal{L}_{ss}$. The overall loss function can be written as:
\begin{equation}
\setlength{\abovedisplayskip}{3pt}
\setlength{\belowdisplayskip}{3pt}
    \mathcal{L}=\mathcal{L}_{rd}+\mathcal{L}_{sc}+\mathcal{L}_{ss}
\end{equation}
These losses jointly regulate the network learning in terms of RRV, temporal, and spatial coherence respectively.

\noindent\textbf{Radial Displacement Loss.} As discussed in Section~\ref{rrv}, the product of RRV measurement and time interval $\Delta t$ can be seen as an approximation of the radial projection of the truth flow vector. This insight is crucial to our self-supervised learning framework because we can use the existing radar inputs to constrain the radial component of flow vectors. Formally, we formulate a radial displacement loss based on Eq.~\ref{velo}:

\begin{equation}\label{rdloss}
\setlength{\abovedisplayskip}{3pt}
\setlength{\belowdisplayskip}{3pt}
\mathcal{L}_{rd}=\sum_{x_i\in\mathcal{P}}|{s}_i^\top \frac{x_i}{||x_i||}-v_i^r\Delta t|
\end{equation}

Despite some inevitable measurement errors, we empirically found that RRV renders strong supervision signals for training scene flow estimators, as shown in Section~\ref{abl:loss}.

\noindent\textbf{Soft Chamfer Loss.}
Introduced by~\cite{wu2020pointpwc}, Chamfer loss is an effective constraint for self-supervised learning of scene flow estimation. By employing the mutual nearest neighbours as pseudo correspondences, it enforces the scene flow to pull the warped source point cloud $\mathcal{P}'=\{\{x'_{i},f_i\}\in\mathbb{R}^{C}\}_{i=1}^{N_1}$ obtained by: $x'_i=x_i+s_i$ as close as possible to the target point cloud $\mathcal{Q}$.

The challenge in our context, however, is that radar point clouds are much sparser and noisier, resulting in quite some points having no real correspondences or even close neighbours in the opposite point cloud. As a result, mapping all points to their nearest neighbours by the vallia Chamfer loss will incur erroneous scene flow estimation. We therefore propose to formulate the Chamfer matching constraints in a probabilistic manner by taking into account the Euclidean distance between a point and its neighbor points in the opposite point cloud. 
Motivated by~\cite{wu2019pointconv}, we further use a kernel density estimation (KDE) method to approximate per-point Gaussian density factor $\nu(x'_i)$ as follows:
\begin{equation}
\setlength{\abovedisplayskip}{3pt}
\setlength{\belowdisplayskip}{3pt}
\nu(x'_i)=\frac{1}{N_2}\sum_{y_j\in\mathcal{Q}}\mathcal{N}(y_j; x'_i, 1)
\end{equation}
Before loss computation, we precompute the density factors for $\mathcal{P}'$ and $\mathcal{Q}$ as prior knowledge about the chance of attaining a successful matching of each point. To mitigate the problem caused by unsatisfying point matching, we select points whose density factor is lower than threshold $\delta$ and discard these points as outliers when computing losses. Our soft Chamfer loss can be formulated as:
\begin{equation}\label{sc}
\setlength{\abovedisplayskip}{3pt}
\setlength{\belowdisplayskip}{3pt}
\begin{aligned}
\mathcal{L}_{sc}=&\sum_{x'_i\in\mathcal{P}'}\mathbb{I}(\nu(x'_i)>\delta)[\min_{y_j\in\mathcal{Q}}||x'_i-y_j||^2_2-\epsilon]_{+} \\
&+\sum_{y_i\in\mathcal{Q}}\mathbb{I}(\nu(y_i)>\delta)[\min_{x'_j\in\mathcal{P}'}||y_i-x'_j||^2_2-\epsilon]_{+}
\end{aligned}
\end{equation}
where $\mathbb{I}(\cdot)$ is the indicator function that returns one when the condition is satisfied, and zero otherwise.
Due to the low resolution of radar sensor, a perfect match between two consecutive clouds is nearly impossible. It is reasonable to let the Chamfer matching tolerate small matching discrepancy. Such toleration is implemented via the $[\cdot]_{+}=\mathrm{max}(0,\cdot)$ operator so that matching errors lower than $\epsilon$ are ignored. 

\noindent\textbf{Spatial Smoothness Loss.}
It has been found that estimating unconstrained point-wise flow vectors can easily lead to ill-posed results~\cite{gojcic2021weakly}. 
For example, two points from the same vehicle might move in different directions or magnitudes. To avoid unreasonable results, it is non-trivial to impose spatial coherence on the estimated motion fields. In~\cite{wu2020pointpwc, kittenplon2021flowstep3d}, a regularization term is added to constrain the predictions when training networks. They formulate a loss function by enforcing the scene flow of $x_i$ close to that of points in its neighbour set $\mathcal{O}(x_i)$.
However, this loss is not suitable to be directly applied to radar point clouds as it implicitly assumes that all points in the neighbour set are close enough to the target point and thus can be considered as measurements from the same rigid body. This assumption does not necessarily hold when it comes to the radar point clouds due to the point sparsity and low resolution as shown in Fig.~\ref{fig:open}.  

To address this challenge, we propose to weigh each neighbour point according to its Euclidean distance to the target point. Intuitively, farther neighbour points are less likely to correlate with the target point and less reliable when enforcing smoothness constraints. To smoothly measure the correlation between point $x_i$ and its neighbour $x_j\in\mathcal{O}(x_i)$, we use the same RBF kernel as in~\cite{pontes2020scene} to formulate the weight: 
\begin{equation}
\setlength{\abovedisplayskip}{3pt}
\setlength{\belowdisplayskip}{3pt}
	k(x_i,x_j)=\exp(\frac{-||x_i-x_j||^2_2}{\alpha})
\end{equation}
where $\alpha$ controls the impact of the distance on the weight. All weight values are normalized together using softmax function. With the assigned weight for each pair of point and neighbour, our spatial correlation loss is formulated as:
\begin{equation}\label{ssloss}
\setlength{\abovedisplayskip}{3pt}
\setlength{\belowdisplayskip}{3pt}
\mathcal{L}_{ss} =\sum_{x_i\in\mathcal{P}}\sum_{x_j\in\mathcal{O}(x_i)}k(x_i,x_j)||{s}_i-{s}_j||^2_2
\end{equation}
By placing local constraints weighted by the metric distance in Eq.~\ref{ssloss}, the spatial coherence of predicted scene flow can be enforced more appropriately for sparser radar point clouds. 



\section{Evaluation}\label{exp}

\subsection{Experiment Settings}

\begin{table}[t]
    \renewcommand\arraystretch{1}
    \setlength\tabcolsep{3.5pt}
    \centering
    \caption{Details of collected sequences for our experiments.}
    \label{seqs}
    \begin{tabular}{c|ccccccc}
    \toprule
       \bf{Sequence}&\it{Seq0}&\it{Seq1}&\it{Seq2}&\it{Seq3}&\it{Seq4}&
       \it{Seq5}&\it{Seq6}\\
    \midrule
       \bf{Period (min)}&12.7&10.0&10.0&10.0&10.0&10.0&10.0\\
       \bf{Distance (km)}&9.5&6.3&5.6&5.7&5.4&5.4&5.2  \\
       \bf{Avg. Speed (km/h)}&44.9&37.8&33.6&34.2&32.4&32.4&31.2\\
       \bf{Usage}&Test&Train&Train&Train&Train&Train&Val\\
    \bottomrule
    \end{tabular}
\end{table}

\begin{table*}[t]
    \renewcommand\arraystretch{1}
    \setlength\tabcolsep{1.5pt}
    \centering
    \caption{Performance of \sysname, \sysname~(without SFR module), and baselines on the test set. } 
    \label{baselines}
    \begin{tabular}{cccccccccc}
    \toprule
       Type & Method & ~ Avg. EPE [m]$\downarrow$ ~ & ~ Avg. RNE [m]$\downarrow$ ~ & 50-50 RNE [m]$\downarrow$ &  Mov. RNE [m]$\downarrow$ & Stat. RNE [m]$\downarrow$ & ~~ SAS$\uparrow$~~ & ~ RAS$\uparrow$ \\
    \midrule
       \multirow{2}*{Non-learning-based}
       & ICP~\cite{121791} & 1.152 & 0.485 & 0.419 & 0.355 & 0.483 & 0.308 & 0.479 \\
       & GraphL-N~\cite{pontes2020scene} & 1.229 & 0.518 & 0.339 & 0.150 & 0.529 & 0.177 & 0.206  \\
    \midrule
       \multirow{7}*{\makecell[c]{Self-supervised \\ learning}}
       & JGWTF~\cite{mittal2020just} & 1.345 & 0.566 & 0.379 & 0.181 & 0.578 & 0.095 & 0.185   \\
       & PointPWC-Net~\cite{wu2020pointpwc} & 1.242 & 0.523 & 0.380 & 0.227 & 0.533 & 0.108 & 0.390 \\
       & GraphL-L~\cite{pontes2020scene} & 1.275 & 0.537 & 0.352 & 0.154 & 0.549 & 0.153 & 0.216  \\
       & FlowStep3D~\cite{kittenplon2021flowstep3d} & 2.111 & 0.889 & 0.670 & 0.439 & 0.900 & 0.081 & 0.242 \\
       & SLIM~\cite{baur2021slim} & 1.224 & 0.515 & 0.338 & 0.149 & 0.527 & 0.178 & 0.207\\
       & {RaFlow}~(w.o. SFR) & {0.608} & {0.256} & {0.185} & \textbf{0.110} & {0.261} & {0.290} & {0.640} \\
       & \textbf{RaFlow} & \textbf{0.570}  & \textbf{0.240}  & \textbf{0.180} & {0.117}  & \textbf{0.244} & \textbf{0.316} & \textbf{0.675}  \\
    \bottomrule
    \end{tabular}
    \vspace{-1.5em}
\end{table*}

\begin{table}[t]
    \renewcommand\arraystretch{1}
    \setlength\tabcolsep{0.5pt}
    \centering
    \caption{Values of all hyperparameters for reproduction.}
    \label{hyper}
    \begin{tabular}{ccc}
    \toprule
       \bf{Parameter}&\bf{Meaning}&\bf{Value}\\
    \midrule
       ${N}_{sc}$& The number of feature extraction scales & 4\\
       ${C}_k, {C}'_k$& The number of local feature channels & 64\\
       ${C}_{cor}$& The number of correlated feature channels & 512\\
       $R$& The radius set for encoding & $\{2,2^2,2^3,2^4\}\mathrm{m}$ \\
       $\zeta$& The threshold to classify points & $0.15$ \\
        $\delta$& The threshold to discard outliers & $0.005$ \\
        $\epsilon$& The threshold for error toleration & $0.1\mathrm{m}$ \\
        $\alpha$& The factor to control smoothness weights & $0.5$ \\
        $N_{nb}$& The number of neighbours for smoothness & $8$ \\
    \bottomrule
    \end{tabular}
\end{table}

\noindent\textbf{Data Collection.} 
An in-house multi-modal dataset was collected to facilitate our evaluation. The collection vehicle was equipped with a set of sensors including a Full-HD RGB camera, an Asensing INS570D navigation system, an RS-Ruby RoboSense (128-beam) LiDAR system and a HASCO LRR30 4D mmWave automotive radar. In particular, the 4D automotive radar has a maximum range of 75m with an azimuth/elevation FOV angle of 120\textdegree/20\textdegree. The range resolution of this 4D radar is $0.2\rm{m}$ while the angular resolution is $\{1.6\degree, 1\degree\}$ for azimuth/elevation measurement, respectively. 
We manually drove the vehicle for over 70 minutes on multiple roads and 7 sequences of data were collected with a total distance of 43.6km. We synchronize radar point clouds to 10Hz LiDAR point clouds and use pose data for motion compensation. For experiments, we select one sequence for test, another one for validation and the rest five for training. Details about these sequences are shown in Table~\ref{seqs}.

\noindent \textbf{Labels.} We use unlabelled samples from the training set for self-supervised learning of DNNs and from the validation set for model selection. 
As the test set is in a moderate amount of data, 
we manually label the test radar frames by referencing a 3D object detection module from the co-located RoboSense LiDAR and the accurate ego-vehicle pose information provided by the RTK-GPS and IMU sensors.
Specifically, the scene flow labels of these samples can be annotated using converted vehicle pose for static points and by tracking bounding boxes (after manual inspection and correction) of dynamic objects for moving points.

\noindent\textbf{Baselines.}
As there is no prior work studying the sparse radar scene flow, our competing approaches are drawn from the general point-based scene flow estimation methods, including \emph{five} state-of-the-art self-supervised 
learning methods: \emph{Just go with the Flow} (JGWTF)~\cite{mittal2020just}, PointPWC-Net~\cite{wu2020pointpwc}, GraphL-L ({the learning version})~\cite{pontes2020scene}, FlowStep3D~\cite{kittenplon2021flowstep3d}, SLIM~\cite{baur2021slim}, and \emph{two} non-learning-based methods: \highlight{Iterative Closest Point (ICP)}~\cite{121791} and GraphL-N ({the non-learning version})~\cite{pontes2020scene}.

\noindent\textbf{Evaluation Metrics.} Prior point-based scene flow estimation works commonly use end point error (EPE) $\mathcal{E}_{i}=||s_{i}-s_{gt,i}||_2$ as the major metric for evaluation. Many of them~\cite{liu2019flownet3d,wu2020pointpwc,pontes2020scene,mittal2020just} achieve a mean EPE lower than $0.2\mathrm{m}$ with dense point clouds (e.g., LiDARs) as input, as shown in their experiments. However, directly applying EPE to radar point clouds lacks sensor-specific consideration and may incur short-sighted conclusion because 4D radars currently have a much lower-resolution than LiDARs. For example, our used 4D radar has a range resolution of $0.2\mathrm{m}$, making a sub-resolution EPE lower than $0.2\mathrm{m}$ nearly impossible. For this reason, we also introduce a metric called Resolution-normalized EPE (RNE) in this work to make our evaluation on par with the ones in LiDAR scene flow estimation. To compensate the resolution gap, we divide the computed EPE by ${\Delta x_i^{r}}/{\Delta x_i^{l}}$, which is the ratio between our radar resolution $\Delta x_i^{r}$ and a common Velodyne LiDAR resolution $\Delta x_i^{l}$ used by previous LiDAR scene flow works~\cite{mittal2020just, wu2020pointpwc, baur2021slim, kittenplon2021flowstep3d, pontes2020scene}.
As both LiDAR~\cite{geiger2013vision} and radar point clouds are generated in the spherical coordinate, each point should have a unique resolution $\Delta x_i$ in the Cartesian coordinate. To implement fine-grained RNE, given the fixed sensor spherical resolution values $\{\Delta r, \Delta \theta, \Delta \phi\}$, we first approximate per-point Cartesian resolution values $\{\Delta X_i, \Delta Y_i, \Delta Z_i\}$ by
\begin{equation}
    \Delta c_i =\sum_{h\in\{r,\theta,\phi\}}\Big|\frac{\partial{c(r,\theta,\phi)}}{\partial{h}}|_{(r,\theta,\phi)= (r_i,\theta_i,\phi_i)} \Big|\Delta h
\end{equation}
where $c\in\{X,Y,Z\}$ denotes the Cartesian coordinate values. We then obtain the point-level resolution $\Delta x_i$ as:
\begin{equation}
    \Delta x_i = \sqrt{(\Delta X_i)^2+(\Delta Y_i)^2+(\Delta Z_i)^2}
\end{equation}
We first compute per-point ratio between $\Delta x_i^r$ and $\Delta x_i^l$ and then obtain the RNE value by dividing EPE by the ratio. 

Following the accuracy score metrics in~\cite{liu2019flownet3d,kittenplon2021flowstep3d,baur2021slim}, we also define two extra metrics based on RNE by calculating the percentage of points that satisfy certain requirements: a) strict accuracy score (SAS): the ratio of points with RNE $\leq$ $0.1\mathrm{m}$ or $10\%$; b) relaxed accuracy score (RAS): the ratio of points with RNE $\leq 0.2\mathrm{m}$ or $20\%$. 
Following~\cite{baur2021slim}, we also report results for static and moving points separately and calculate their average to avoid the imbalance of these two classes. For notation, we use 50-50 RNE as the average of between Stat. RNE and Mov. RNE, while Avg. RNE denotes the mean RNE for all test points.

\noindent\textbf{Training Details.} We trained our model for 50 epochs using the Adam optimizer. The initial learning rate is set as 0.001 and decays by 0.9 after each epoch. Data augmentation is implemented by randomly rotating and translating each point cloud from the training set. During training, we downsample all point clouds to $N_1=N_2=256$ for fast batch processing but do not downsample test samples in order to estimate the flow vector of each point. The values of hyperparameters can be seen in Table~\ref{hyper}, which are fixed for all experiments.

\subsection{Quantitative Results}\label{quantitative}
Our experiments begin with the comparison among \sysname~and baselines. For a fair comparison, we train networks of self-supervised baselines~\cite{mittal2020just,wu2020pointpwc,pontes2020scene,kittenplon2021flowstep3d,baur2021slim} under the same setting as ours. For the evaluation of ICP~\cite{121791}, we iteratively optimize its rigid ego-motion output and apply this transformation to all points to generate scene flow. We also iteratively optimize the objective function of GraphL-N~\cite{pontes2020scene} to obtain the final scene flow prediction. 

As seen in Table~\ref{baselines}, \sysname~outperforms other self-supervised learning and non-learning-based methods on all metrics. This confirms the effectiveness of our radar-oriented architecture design and specific loss functions to cope with sparse, noisy and low-resolution radar point clouds. 

It is also worth noting the importance of the SFR module that refines all static points through a predicted rigid transformation. As we can see, while \sysname~ without SFR can still achieve comparable results, the Stat. RNE increases by $6.5\%$ and the SAS decreases by $9.0\%$ due to the absence of static flow regularization in the process. We also observe that the Mov. RNE decreases by 0.7$\rm{cm}$ without the SFR module. \highlight{This drop is reasonable since the SFR module cannot be perfect and it inevitably mis-classifies a few moving points to static, whose flow vectors are further replaced by the rigid flow derived from the estimated rigid ego-motion. In our data collection, moving points (e.g., vehicles and motorcycles etc.) are relatively fewer than the static points (e.g., trees and road railings etc.), which can be seen in Fig.~\ref{fig:qual}. It is therefore a trade-off to achieve the optimal RNE on average by sacrificing a little RNE performance (-0.007m) of dynamic points for a large improvement margin (+0.017m) of static points. If the performance on dynamic points is favored by the user, the hyperparameter $\zeta$ can tuned to be smaller during inference.} 

Despite the satisfying performance achieved on dense point clouds (e.g., LiDAR), state-of-the-art methods all struggle in delivering the same efficacy on radar data. Interestingly, the traditional ICP~\cite{121791} achieves better results than other baselines on Avg. RNE because it can solve accurate static flow vectors in many simple scenarios (e.g., ego-vehicle is stationary). Nevertheless, ICP falls behind on 50-50 RNE due to its incompatibility to tackle moving points. 

\vspace{-1em}
\subsection{Qualitative Results}
\begin{figure}[t]
    \centerline{\includegraphics[scale=0.243]{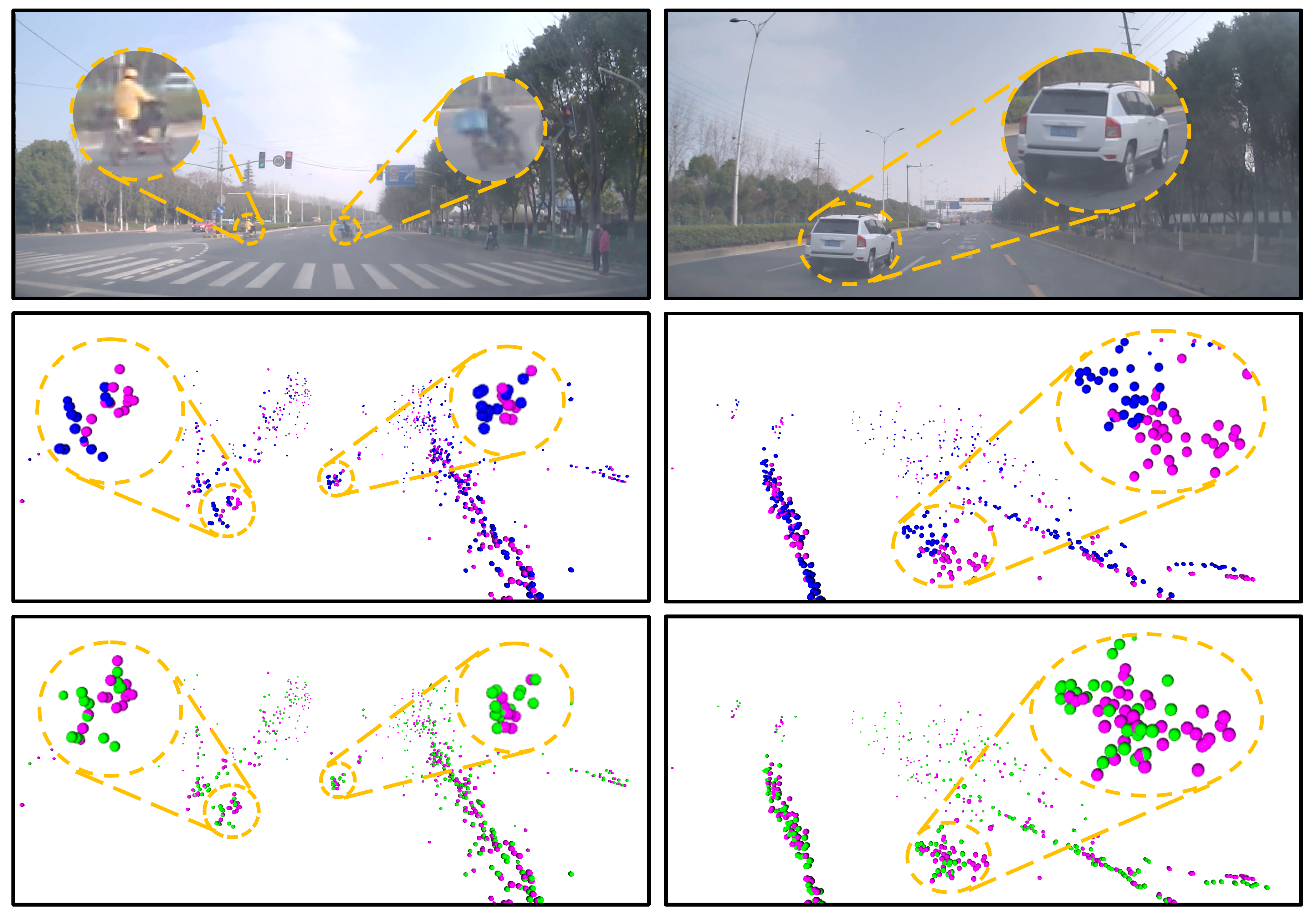}}
    \caption{Scene flow estimation visualization. Figures on the top are the corresponding images captured by camera. The middle row shows points from $\mathcal{P}$ (\textcolor[rgb]{0,0,1}{blue}) and $\mathcal{Q}$ (\textcolor[rgb]{1,0,1}{magenta}) respectively, while the bottom figures are the warped point cloud $\mathcal{P}'$ (\textcolor[rgb]{0,1,0}{green}) and point cloud $\mathcal{Q}$. \textcolor[rgb]{1,0.75,0}{Yellow} circles denote zooming-in operation. It can be seen that, after applying
    the predicted scene flow, the warped points (\textcolor[rgb]{0,1,0}{green}) are clearly closer to the (\textcolor[rgb]{1,0,1}{magenta}) points in
    the next frame, either for the moving vehicles, motorcycles or the static road railing. This point
    alignment across frames demonstrates the accurate radar scene flow estimation. More qualitative results can be seen in our submitted video. (Best viewed in color) }
    \label{fig:qual}
\end{figure}

To intuitively show the performance of \sysname, we follow~\cite{liu2019flownet3d} and visualize our scene flow estimation results by warping point cloud $\mathcal{P}$ with the scene flow in Figure~\ref{fig:qual}. In the scene described by the left column, our ego-vehicle is stationary while two highlighted motorcycles are moving forward. In the right scene, one cars are driving forward at a slower speed as ours. It is clear that our method can correctly estimate the flow vectors of both static and moving points whether the ego-vehicle is moving or keeps still.

\vspace{-1em}
\subsection{Ablation Study}

\noindent\textbf{Impact of Features.}\label{abl:ft}
Besides the 3D positional information, we use the RRV, RCS and Power measurements as auxiliary input features which are unique strengths of radar. To understand their impacts, we ablate each component and report the performance change. As shown in Table~\ref{features}, removing the RRV feature has the largest impact on our method. We attribute this to the fact, that under the supervision of our radial replacement loss, our model can implicitly learn to exploit the motion cues provided by the RRV feature (c.f. Sec.~\ref{loss}). 
The RCS and Power features can also make non-negligible contributions to our method because of the extra semantic information provided by them. 

\begin{table}[t]
    \renewcommand\arraystretch{1}
    \setlength\tabcolsep{8pt}
    \centering
    \caption{Ablation results of input features.}
    \label{features}
    \begin{tabular}{cccccc}
    \toprule
       {RRV}&{RCS}&{Power}&{Avg. RNE [m]}$\downarrow$&{SAS}$\uparrow$&{RAS}$\uparrow$\\
    \midrule
      $\checkmark$& $\checkmark$ & & 0.247 & 0.305 & 0.657 \\
      $\checkmark$& &$\checkmark$ & 0.246 & 0.302 & 0.658   \\
      & $\checkmark$&$\checkmark$ & 0.362 & 0.204 & 0.419  \\
      $\checkmark$&$\checkmark$&$\checkmark$& \textbf{0.240}&\textbf{0.316}&\textbf{0.675} \\ 
    \bottomrule
    \end{tabular}
\end{table}

\vspace{-1em}
\noindent\textbf{Impact of Losses.}\label{abl:loss}
To validate the individual effectiveness of our three specific loss functions, we also conduct ablation studies with different combinations of them. Moreover, we also include the previous Chamfer loss $\mathcal{L}_c$ and smoothness term $\mathcal{L}_s$ used by~\cite{wu2020pointpwc,kittenplon2021flowstep3d} as reference for our loss functions. The results can be seen in Table~\ref{loss table}. The bottom combination is our \sysname with full loss functions. It is clear that each loss term improves the overall performance. Specifically, the radial displacement loss $\mathcal{L}_{rd}$ counts for more than another two losses due to its strong supervision in the radial direction. When replacing the previous \emph{hard} Chamfer loss $\mathcal{L}_{c}$ with our soft Chamfer loss $\mathcal{L}_{sc}$, the results are upgraded by a large margin. This confirms the effectiveness of our proposed a) outlier discarding scheme that loosens the strong assumption imposed by absolute bijective mapping and b) small error toleration design where only errors originated from wrong correspondence matching are considered for network update. Last but not least, our spatial smoothness loss $\mathcal{L}_{ss}$ also shows advantages over the prior smoothness term $\mathcal{L}_s$. 

\begin{table}[t]
    \renewcommand\arraystretch{1}
    \setlength\tabcolsep{10pt}
    \centering
    \caption{Ablation results of loss functions.}
    \label{loss table}
    \begin{tabular}{cccc}
    \toprule
       Losses&{Avg. RNE [m]}$\downarrow$&{SAS}$\uparrow$&{RAS}$\uparrow$\\
    \midrule
       $\mathcal{L}_{c}+\mathcal{L}_{ss}+\mathcal{L}_{rd}$&0.450&0.165&0.421\\
       $\mathcal{L}_{sc}+\mathcal{L}_{s}+\mathcal{L}_{rd}$&0.242&0.308&0.675 \\
       $\mathcal{L}_{sc}+\mathcal{L}_{rd}$&0.255&0.296&0.652\\
       $\mathcal{L}_{ss}+\mathcal{L}_{rd}$&0.257&0.281&0.635\\
       $\mathcal{L}_{sc}+\mathcal{L}_{ss}$&0.405&0.236&0.375\\
       $\mathcal{L}_{sc}+\mathcal{L}_{ss}+\mathcal{L}_{rd}$&\textbf{0.240}&\textbf{0.316}&\textbf{0.675}\\ 
    \bottomrule
    \end{tabular}
\end{table}

\vspace{-1em}
\subsection{Downstream Motion Segmentation Task}

\begin{figure}[t]
    \centerline{\includegraphics[scale=0.225]{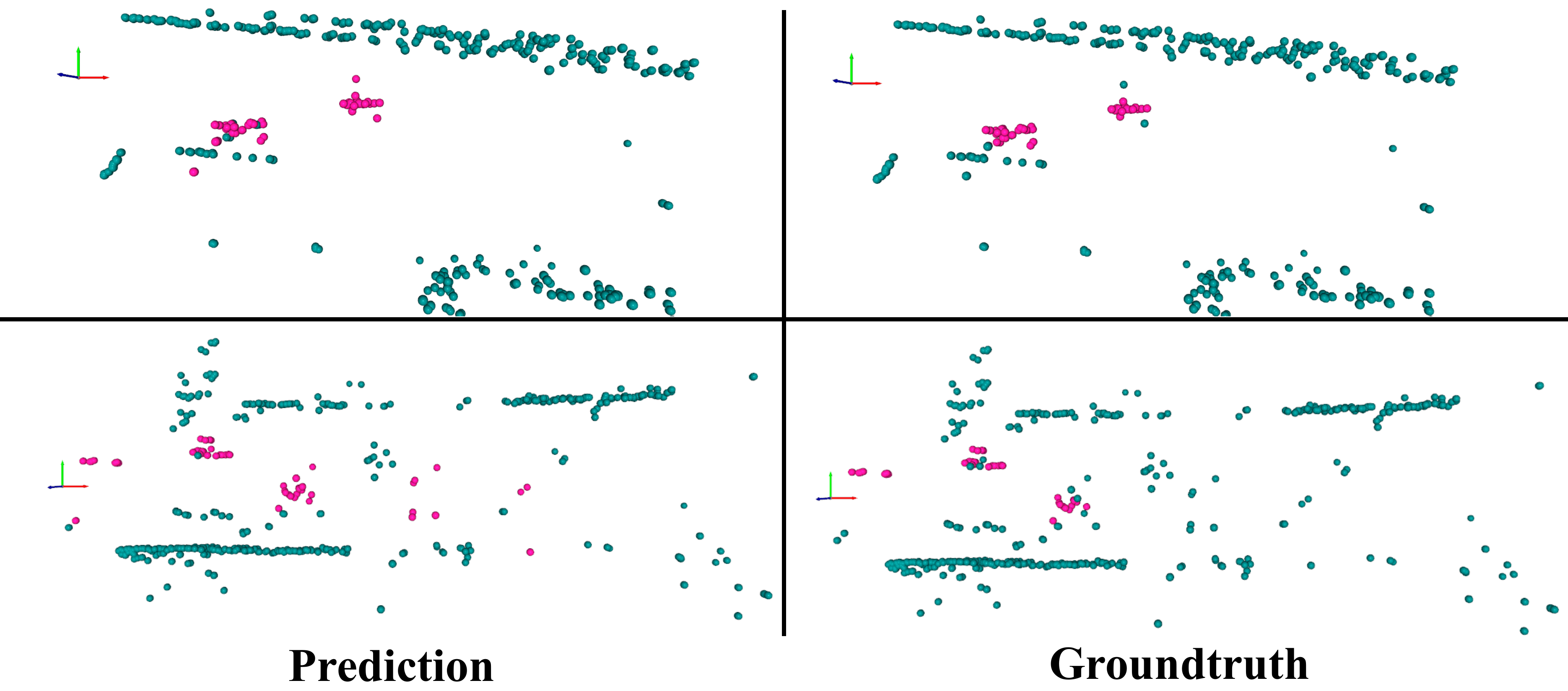}}
    \caption{Visualization of motion segmentation results. Left column shows our prediction while right column is the ground truth. {Moving} and {stationary} points are rendered as \textcolor[rgb]{1,0.08,0.58}{pink} and \textcolor[rgb]{0,0.5,0.5}{teal}, respectively. Note that this is non-trivial as the ego-vehicle is also \emph{moving} in both two scenes. (Best viewed in color)} 
    \label{motion}
    \vspace{1em}
\end{figure}

As one of the outputs of our scene flow estimation pipeline, the generated static mask (c.f. Sec.~\ref{refinement}) 
can be used to segment scene points into stationary and moving points. Following the metrics used by~\cite{baur2021slim}, we evaluate our model on the test set for the motion segmentation task. As a result, an accuracy score of $81.9\%$ can be achieved with a mIoU score of $47.9\%$ and a sensitivity of $82.7\%$. Qualitative results for two scenes can be seen in Figure~\ref{motion}. 

\section{Limitation and Future Work}
\highlight{As the first attempt of radar scene flow estimation, our method has room to improve. First, our performance is still somewhat limited due to the lack of real supervision signals. Better accuracy might be needed in practice to unlock a wider range of downstream tasks, e.g., multi-object tracking and point cloud stitching. To this end, cross-modal supervision signals from other co-located sensors (e.g., RGB camera or IMU) will be exploited in conjunction with our self-supervised learning to further improve the flow estimation accuracy. Second, as we discussed in Sec.~\ref{quantitative}, due to the introduction of the SFR module, the performance on a few moving points is sacrificed to trade for a larger performance gain on average. Considering the importance of the moving points for autonomous driving, more investigation is needed to allow adaptive thresholding of $\zeta$ in response to different road situations on-the-fly. Third, the constant velocity assumption introduced in Sec.~\ref{rrv} might not hold in some cases where the vehicle platform has a large acceleration between two frames. Higher radar sampling rates or short time intervals between frames are needed to mitigate such extreme situations.} 
\section{Conclusion}

We present a self-supervised learning method called \sysname~to estimate scene flow on 4D radar point clouds. A novel architecture and three loss functions are specifically designed to address the challenges induced by the characteristics of radar sensors. We collect a multi-modal dataset by driving a vehicle in the wild and compare \sysname~with state-of-the-art point-based scene flow methods. We validate the effectiveness of our designs in different aspects and show that the result of \sysname can enable downstream motion segmentation tasks. 

\newpage
\bibliographystyle{IEEEtran}
\bibliography{reference}

\vfill

\end{document}